\documentclass[twoside,11pt]{article}
\usepackage{jair, theapa, rawfonts}
\usepackage{times}
\usepackage{latexsym}
\usepackage{url}
\usepackage{arydshln}
\usepackage{graphicx}
\usepackage{amsmath}
\usepackage{amsfonts, amssymb}
\usepackage{xcolor}
\usepackage{multirow}
\usepackage{arydshln}

\jairheading{61}{2018}{1-28}{06/17}{03/18}
\ShortHeadings{Corpus-Level Fine-Grained Entity Typing}
{Yaghoobzadeh, Adel, \& Sch\"{u}tze}
\firstpageno{1}

\def\dnrm#1{\mbox{$_{\hbox{\scriptsize #1}}$}}
\def\uprm#1{\mbox{$^{\hbox{\scriptsize #1}}$}}

\newcommand{\todooff}{\long\gdef\todo##1{}}
\newcommand{\todoon}{\long\gdef\todo##1{{
\bf\textcolor{red} {TODO: ##1}
}}}
\todoon
\todooff

\newcounter{notecounter}

\newcommand{\enoteson}{\long\gdef\enote##1##2{{
\stepcounter{notecounter}
{\large\bf
\hspace{1cm}\arabic{notecounter} $<<<$ ##1: ##2
$>>>$\hspace{1cm}}}}}
\enoteson

\def\figref#1{Figure~\ref{fig:#1}}
\def\figlabel#1{\label{fig:#1}\label{p:#1}}

\def\tabref#1{Table~\ref{tab:#1}}
\def\tablabel#1{\label{tab:#1}\label{p:#1}}

\def\secref#1{Section~\ref{sec:#1}}
\def\seclabel#1{\label{sec:#1}}
\def\eqref#1{Eq.~\ref{eqn:#1}}

\def\eqlabel#1{\label{eqn:#1}}

\def\gmodel{global model}
\def\cmodel{context model}

\begin{document}

\title{Corpus-Level Fine-Grained Entity Typing}

\author{\name Yadollah Yaghoobzadeh  \email yadollah.yaghoobzadeh@microsoft.com \\
      \addr  Microsoft Research Montreal \\
       Montreal, Canada \& \\
       Center for Information and Language Processing \\
       LMU Munich, Munich, Germany
       \AND
       \name Heike Adel \email heike.adel@ims.uni-stuttgart.de \\
       \addr Institute for Natural Language Processing\\
       University of Stuttgart, Stuttgart, Germany \& \\
       Center for Information and Language Processing \\
       LMU Munich, Munich, Germany
		\AND
		 \name Hinrich  Sch\"{u}tze  \email inquiries@cislmu.org\\
       \addr Center for Information and Language Processing \\
       LMU Munich, Munich, Germany
       }

\maketitle

\begin{abstract}
Extracting information about entities remains an important research area. This paper addresses the problem of corpus-level entity
typing, i.e., inferring from a large corpus that an entity
is a member of a class, such as ``food'' or ``artist''.  
The application of entity typing we are interested in is
knowledge base completion, specifically, to learn which
classes an entity is a member of.
We propose \textsc{figment} to tackle this problem.  
\textsc{Figment} is embedding-based and combines (i) a \gmodel{} that
computes scores based on global information of an
entity and (ii) a \cmodel{} that first evaluates the individual
occurrences of an entity and then aggregates the scores.

Each of the two proposed models has specific properties. 
For the global model, learning high-quality entity representations is crucial because
it is the only source used for the predictions.
Therefore, we introduce representations using the name and  contexts 
of entities on the three levels of entity, word, and character.
We show that each level provides complementary information and a multi-level
representation performs best.
For the context model, we need to use distant supervision since 
there are no context-level labels available for entities. 
Distantly supervised labels are noisy and this harms the performance of models. 
Therefore, we introduce and apply new algorithms for noise mitigation using 
multi-instance learning.
We show the effectiveness of our models on a large entity typing dataset
built from Freebase.

\end{abstract}

\section{Introduction}
\label{Introduction}

Knowledge about entities is essential for natural language understanding (NLU).
This knowledge includes facts about entities, such as their names, properties, relations and types.
This data is usually stored in large-scale structures called knowledge bases (KBs)  and therefore building and maintaining KBs is very important.
Examples of such KBs are  Freebase \cite{bollacker2008freebase}, YAGO \cite{suchanek2007yago} and Wikidata \cite{wikidata} . 
KB structure is
usually equivalent to a graph in which entities are nodes
and edges are relations between entities. 
Each node is also associated with one or more semantic
classes, called types.

Incompleteness is a crucial challenge for KBs 
because the world is changing -- new entities are emerging, and existing entities are getting new properties.
Therefore, there is always the need to update KBs. 
To do so, we propose an information extraction method that processes large raw corpora in order to gather knowledge about entities.
Most prior work tries to complete relations between entities.
In contrast, the focus of this work is on the completion of entity types in KBs.
For example, given a large corpus and the entity ``Barack Obama'' we need to find all his types including ``person'', ``{politician}'', and ``{author}''.

We define our problem as follows.
Let $K$ be a knowledge base that models a
set $E$ of entities, a set $T$ of fine-grained classes or \emph{types}
and a membership function $m:  E \times T \mapsto \{0,1\}$ such that
$m(e,t)=1$ iff entity $e$ has type $t$. Let $C$ be a large
corpus of text. Then, the problem we address in this paper
is \emph{corpus-level fine-grained entity typing}:
For a given pair of entity $e$ and type $t$ determine -- based
on the evidence available in $C$ -- whether 
$e$ is a member of type $t$ (i.e., $m(e,t)=1$)
or not (i.e., $m(e,t)=0$)
and update the membership relation $m$
of $K$ with this information.
We investigate two approaches to entity typing:
a \gmodel{} and a \cmodel{}.

The \textbf{\gmodel} aggregates all information about an
entity from the corpus and its name, and then 
predicts the type
based on that. 
We represent necessary information about the entity in a multi-level
representation. 
This representation is global, meaning that it does not describe a specific context or mention of an entity but rather the aggregated information about it.
One important source to compute the entity representations is the
\emph{contexts in which the entity is used}.  We take
the standard method of learning embeddings for words and
extend it to learning embeddings for entities. This requires
the use of an entity linker and can be implemented by
replacing all occurrences of the entity by a unique token (e.g., \citeR{Wang14joint,figment15,figment17}).
We refer to entity embeddings as \emph{entity-level
  representations}. 
Entity-level representations are often uninformative for
rare entities, so relying only on entity embeddings
is likely to produce poor results. 
In this paper, we use
\emph{entity names}
as a source of information that is
complementary to entity embeddings. 
We define an entity name as a 
noun phrase that is used to refer to an entity.
We learn character-level and word-level representations of 
entity names.  

For the \emph{character-level representation}, 
we adopt different character-level 
neural
network architectures. 
Our intuition is that there is
sub/cross-word information (e.g., orthographic patterns) that is
helpful to get better entity representations, especially for
rare entities. A simple example is that a three-token
sequence containing an initial like ``P.'' surrounded by two
capitalized words (``Rolph P.\ Kugl'') is likely to refer to a person.

We compute
the \emph{word-level representation} as the sum of the
embeddings of the words that make up the entity name.
The sum of
the embeddings accumulates evidence for a
type/property over all constituents, e.g., 
a name
containing  ``stadium'', ``lake'' or
``cemetery'' is likely to refer to a location. 
In this paper, we compute our word-level representation 
with two types of word embeddings:
(i) using only contextual information of words in the corpus, e.g., 
by \textsc{word2vec} \cite{mikolov2013efficient}
and
(ii) using subwords as well as contextual information of words,
e.g., by Facebook's recently released  \textsc{fasttext} \cite{subword16}.

The \textbf{\cmodel{}} first scores each individual context of $e$
as expressing type $t$ or not.
The final value 
of $m(e,t)$ is then computed based on the distribution of
context scores.  
One difficulty of this model is that 
it is too expensive to label
each entity context with its labels.  
Distant supervision \cite{Mintz09} is used to  reduce the need for manually labelling 
contexts.
Distant supervision assumes that if 
an entity has a type in the KB, 
then all contexts mentioning that entity 
express that type. 
However, this assumption is too
strong and gives rise to many \emph{noisy} labels.
Different techniques
to deal with that problem 
have been investigated.
The main technique  is multi-instance (MI) learning \cite{riedel2010}. 
It relaxes the distant supervision assumption to the assumption
that at least one instance of a bag (collection of 
all sentences containing the given entity) 
expresses the type given in the KB.
Multi-instance multi-label (MIML) learning 
is a generalization of MI in which one bag
can have several labels \cite{surdeanu2012multi}. 

Most MI and MIML methods are based on hand-crafted features.
Recently, \citeA{Zeng15emnlp} introduced an end-to-end
approach to
MI learning based on neural networks.
Their MI method takes the most confident instance
as the prediction of the bag.
\citeA{lin2016} further improved that method  by taking
other instances into account as well;
they proposed MI learning based on
selective attention as an alternative way of
relaxing the impact of noisy labels on relation extraction. 
In selective attention, a weighted average of instance
representations is calculated first and then used 
to compute the prediction of a bag.

In this paper, we introduce four multi-label versions of 
MI.
(i) \emph{MIML-MAX}
takes the maximum of instance scores
for each label. 
(ii) \emph{MIML-MEAN} takes the mean of instance scores
for each label. 
(iii) \emph{MIML-MAX-MEAN} takes the maximum and the mean of instance scores 
in training and testing, respectively.
(iv) \emph{MIML-ATT} applies, for each label, selective attention 
to the instances.
We apply these MIML algorithms to fine-grained entity typing
and show that 
MIML-ATT 
deals well with noise in
this task and 
gives clear improvements for the distantly supervised models.

The \gmodel{} is potentially more robust  compared to the \cmodel{} since it takes
into account all the available information at once. 
In contrast, the \cmodel{} has the advantage that it can correctly predict types for
which there is only a small number of reliable
contexts. For example, in a large corpus, it is likely to
find a few reliable contexts indicating that ``Barack Obama'' is
a best-selling author even though this evidence 
may be obscured in the global distribution
because the vast majority of mentions of ``Obama'' does not occur
in author contexts.

We implement the \gmodel{} and the \cmodel{} as well
as a simple combination of the two and call the resulting system
\textsc{figment}:
FIne-Grained eMbedding-based ENtity Typing.
A key feature
of \textsc{figment} is that it makes extensive use of 
distributed vector representations
or \emph{embeddings}.

The contributions in this paper include the following:
(i) We address fine-grained entity typing by using text corpora for the application in knowledge base completion.
(ii) We build a dataset for this task from Freebase entities and their fine-grained types.
(iii) We introduce, implement and compare two types of models for the task, global and context models,  and a joint model of them.
(iv) We represent entities using novel distributed
representations on the three levels of entity, word and character. 
(v) We introduce new algorithms for multi-instance learning in neural networks and apply them for the first time to the task of fine-grained entity typing.

\section{Related Work}
\seclabel{relatedwork}
Our task is \emph{fine-grained entity typing}.
\citeA{neelakantan2015inferring}, \citeA{Suzuki2016et} and \citeA{xie16dkrl} also address a similar task,
but they rely on entity descriptions in KBs.
Thus, in contrast to our approach, their system is not able to type entities that are not
covered by existing KBs.
We infer
classes for entities from a large corpus and do not assume
that these entities occur in the KB.
The problem of \emph{fine-grained mention typing} (FGMT) 
\cite{spaniol2012hyena,ling2012fine,yogatama2015acl,delcorro15finet,attentiveTyper16,partialLabel16,RabinovichK17,erpv17multilingualFGEM,shimaoka17eacl} 
is related to our task.
FGMT classifies single \emph{mentions} of named entities
to their context-dependent types whereas we attempt to identify all types of a KB \emph{entity} from the aggregation of all its mentions.
FGMT can still be evaluated in our task by aggregating 
the mention-level decisions.

\emph{Entity set expansion} (ESE)
is the problem of finding entities in a class (e.g.,
medications) given a seed set 
 (e.g., 
\{``Ibuprofen'',
``Maalox'', ``Prozac''\}).
The standard solution is pattern-based bootstrapping
\cite{thelen2002bootstrapping,gupta14evalpatterns}.  
ESE is different from the problem we address because ESE
starts with a small seed set whereas we assume that 
a large number of examples from a knowledge base (KB) is available. Initial experiments 
with the system of \citeA{gupta14evalpatterns} show that
it was not performing well for our task -- this is not
surprising given that it is designed for a task with
properties quite different from entity typing.

Fine-grained entity typing can be used for \emph{knowledge base completion (KBC)}.
Most KBC systems focus on \emph{relations} between entities,
not on \emph{types} as we do.
Some  generalize the patterns of relationships within the KB \cite{nickel12yago,bordes2013transe},
while  others
use a combination of within-KB generalization and
information extraction from text
\cite{Weston2013Conn,socher2013reasoning,Jiang12linkpred,Riedel13universal,Wang14joint}.

We also introduce methods for \emph{noise mitigation in distant supervision}.
Distant supervision can be used to train
information extraction systems, e.g., in 
relation extraction (e.g., \citeR{Mintz09,riedel2010,hoffmann2011,Zeng15emnlp,adel2016})
and entity typing (e.g.,
\citeR{ling2012fine,yogatama2015acl,donghybrid}).
To mitigate the noisy label problem, 
multi-instance (MI) learning has been introduced
and applied in relation extraction 
\cite{riedel2010,ritter13tacl}. 
\citeA{surdeanu2012multi} introduce multi-instance multi-label (MIML) learning
to extend MI learning for multi-label relation extraction.
Those models are based on manually designed features.
\citeA{Zeng15emnlp} and \citeA{lin2016} introduce MI 
learning methods for neural networks. 
We introduce MIML algorithms
for neural networks.
In contrast to most MI/MIML methods, which are applied in relation extraction, 
we apply MIML to the task of 
fine-grained entity typing.
\citeA{ritter13tacl} apply MI 
on a Twitter dataset with ten types.
Our dataset has a larger number of classes or types (namely 102)
and input examples, 
compared to that Twitter dataset and also to
the most widely used datasets for evaluating MI (cf., \citeR{riedel2010}).
This makes our setup more challenging 
because of different dependencies
and the multi-label nature of the problem. 
Also, there seems to be a difference in how entity relations and entity types are mentioned: expressing the entity relations is more likely to be explicit than entity types in many text genres. This means we usually need to look at many entity contexts for reliable type predictions. 
This can influence the choice of MIML algorithms since some of them just pick one context (instance) for prediction.
Our experimental results confirm this hypothesis. 

\section{Motivation and Background}
In this section, we provide motivation and  background information
for our work, including Freebase, incompleteness of knowledge bases, entity linking and FIGER types.
\subsection{Freebase}

Large scale knowledge bases (KBs) like Freebase
\cite{bollacker2008freebase}, YAGO \cite{suchanek2007yago}
and Wikidata \cite{wikidata} are designed to store world knowledge.  
KB structure is usually graph-based and with different schemas. 
Here in this work, we use Freebase. 
Freebase is a labeled graph, with nodes and directed edges. 
Topics (or entities) are the essential part of Freebase, which are
represented as graph nodes. 
These topics can be named entities (like ``Germany'') or abstract concepts (like ``love''). 
In this work, we refer to Freebase topics as entities. 
Apart from entities, Freebase uses types like ``{city}'', ``{country}'', ``{book\_subject}'', ``{person}'', etc. 
Each entity can have one or many types, e.g., 
``Arnold Schwarzenegger'' is a ``{person}'', ``{actor}'', ``{politician}'', ``{sports\_figure}'', etc.
There are about 1,500 types in Freebase, organized by domains; e.g.,
the domain ``{food}'' has types like ``{food}'', ``{ingredient}'' and
``{restaurant}''.
Each type contains some specific properties about entities,
e.g., the ``{actor}'' type contains a property that lists all
films that ``Arnold Schwarzenegger'' has acted in.
In other words, entities are connected to each other by properties because they 
are in certain types. 
For example, ``Arnold Schwarzenegger'' is connected to ``California'' with
property ``governor\_of'' which is a property defined for the type ``{politician}''.

\subsection{Incompleteness of Knowledge Base}
Even though Freebase is the largest publicly available
KB of its kind, it still has significant
coverage problems. For example,
78.5\% of persons in Freebase do not have
a \textit{nationality} \cite{min2013distant}, or in our test set,
12\% of persons do not have a finer grained type.

This is unavoidable
partly because Freebase is user-generated and partly because
the world changes constantly and Freebase has to be updated to reflect
those changes.  All existing KBs that attempt to
model a large part of the world suffer from this
incompleteness problem.
Incompleteness is likely to become an even bigger problem in
the future as the number of types covered by KBs increases. As more and more fine-grained types are
added, achieving good coverage for these new
types using only human editors will become impossible.

The approach we adopt in this paper to address
incompleteness of KBs is extraction of information from
large text corpora. Text is arguably the main
repository for the type of knowledge represented in KBs, and thus
it is reasonable to attempt completing them based on
text. There is a significant body of work on
corpus-based methods for extracting knowledge from text.
However, most of the work has addressed relation extraction,
and not the acquisition of type information -- roughly
corresponding to unary relations (see \secref{relatedwork}).  In
this paper, we focus on typing entities.

\subsection{Entity Linking}
The first step in extracting information about entities from
text is to reliably identify mentions of these
entities.
This problem of \emph{entity linking} has some mutual
dependencies with our task, entity typing.
Indeed, some recent work demonstrates large improvements when entity
typing and linking are jointly modeled
\cite{ling15entitylinking,durrett14entityanalysis}.
However, there are
constraints that are important for high-performance entity
linking but are of little relevance to entity typing.
For example, there is large body of literature on entity linking
that deals with 
coreference resolution and
inter-entity constraints --
e.g., ``Naples'' is more likely to refer to a US (resp.\ an
Italian) city in a context mentioning ``Fort Myers''
(resp.\ ``Sicily'').  
Therefore, we only address
entity typing in this paper
and
consider entity linking as an independent module 
that provides contexts of entities for \textsc{figment}.
(A similar problem definition is used in relation extraction, cf., 
\citeR{Zeng15emnlp,lin2016}.)
More specifically, we  build \textsc{figment} on top of the output of an
existing entity linking system and use
FACC1 \cite{gabrilovich2013facc1}, an automatic Freebase annotation of 
ClueWeb \cite{clueweb12url}.
According to the FACC1 distributors,
precision of annotated entities is around 80-85\% and recall
is around 70-85\%.

\subsection{FIGER Types}
Our goal is fine-grained typing of entities. However, there are types 
which are too fine-grained, such as 
``Vietnamese urban district''.
To create a reliable setup for evaluation and to make sure
that all types have a reasonable number of instances, we
adopt the FIGER type set
\cite{ling2012fine} that was created with the same goals in mind.
FIGER consists of 113 tags and was created in an attempt to preserve the diversity of
Freebase types while consolidating infrequent and unusual
types through filtering and merging. For example,
the Freebase types 
``{dish}'', ``{ingredient}'', ``{food}'' and
``{cheese}''
are mapped to one type ``{food}'' (for a complete list of  FIGER types, see \citeR{ling2012fine}.).
 We use ``type'' 
to refer to FIGER types in the rest of the paper.

\section{Global, Context, and Joint Models}
\seclabel{task}
\begin{figure}[ht]
\centering
\includegraphics[scale=0.40]{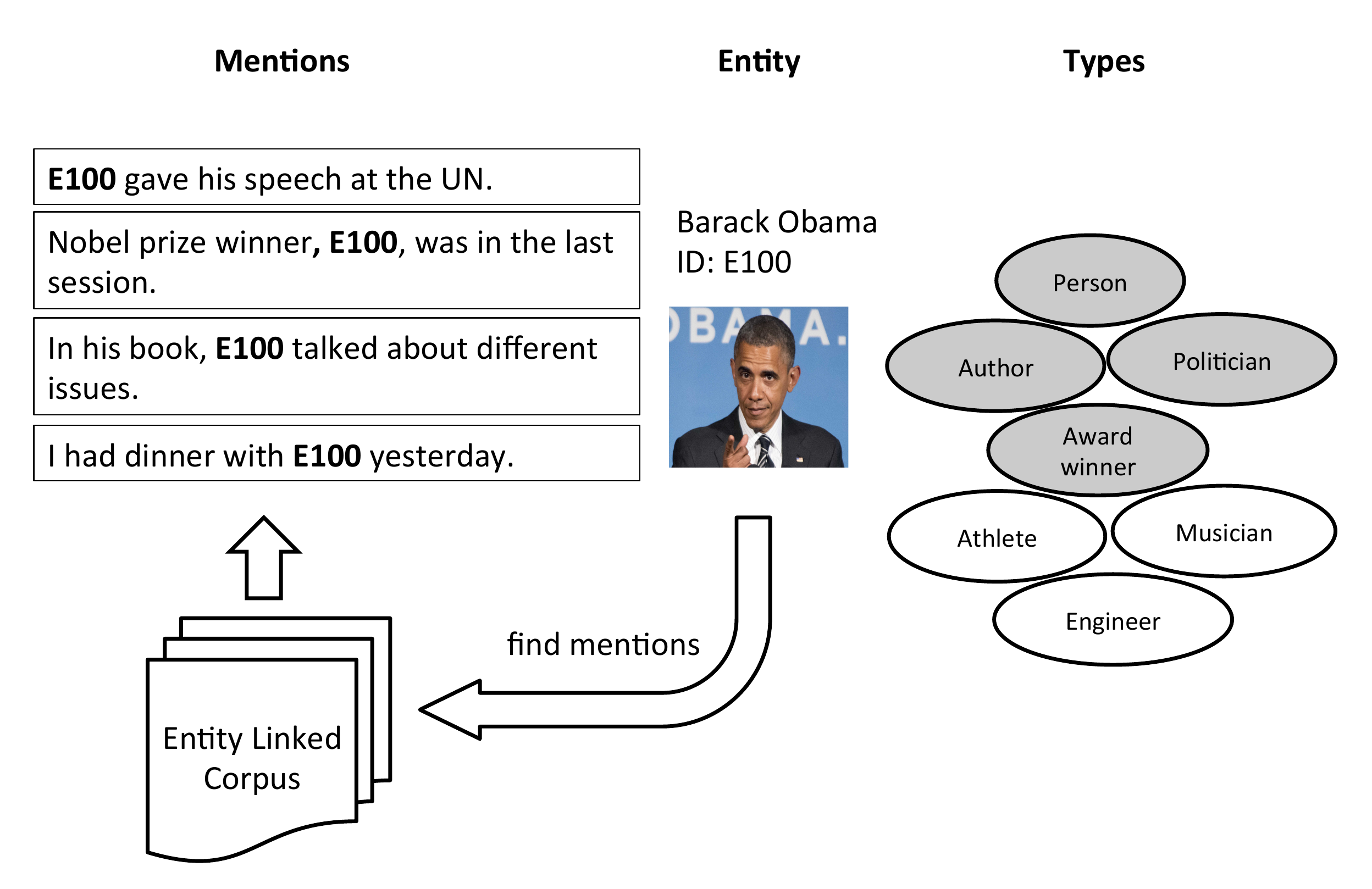}
\caption{An example for corpus-level fine-grained entity typing. 
Given Barack Obama with ID of E100 from a KB, and an entity linked corpus with some mentions of E100, the task is to predict the correct fine-grained types of E100 : ``{person}'', ``{politician}'', and ``{author}'', and ``{award-winner}''.}
\figlabel{task}
\end{figure}

Given (i)
a KB with a set of entities $E$, (ii) a set of types $T$,
and (iii) a large corpus $C$ in which mentions of 
$E$ are
linked, we 
address the 
task of \emph{corpus-level fine-grained entity typing}: 
predicting whether  
entity $e$ is a member of type $t$ or not.
We show an example diagram of our task in \figref{task}.
We use a set of training examples
to learn $P(t|e)$:
the probability that entity $e$ has type $t$.
These probabilities  can be used to assign \emph{new types} to
entities covered in the KB as well as typing \emph{new} or \emph{unknown entities} -- i.e., entities not
covered by the KB.
For new or unknown entities,
an entity linking system would be necessary that
identifies and clusters entity mentions. Examples for
those systems are the ones participating in TAC KBP \cite{mcnamee2009overview}.

In this section, we introduce our two models, global and context, 
as well as their combination (joint model).
In \tabref{overview}, we provide a brief overview
of the models. 
Details are given in the following 
sections.

\textbf{Notations and definitions.}
Lowercase letters (e.g., $e$) refer to variables.
Bold lowercase letters  (e.g., $\mathbf{e}$) are vectors and
bold uppercase letter (e.g., $\mathbf{W}$) are matrices.
We define $\mbox{BCE}$, the binary cross entropy between two variables,
as follows where $y$ is a binary variable and 
$\hat{y}$ is a real valued variable between 0 and 1.
\begin{equation}
\mbox{BCE}(y, \hat{y}) = - \Big(y \log(\hat{y}) + (1 - y) (1 - \log(\hat{y})) \Big)\nonumber
\end{equation}

\begin{table}[h]
\centering
\footnotesize{
\setlength{\tabcolsep}{2pt}{
\begin{tabular}{p{6cm} | p{6cm}}
\multicolumn{1}{c}{Global Model} & \multicolumn{1}{c}{Context Model} 
 \\
\hline 
$\begin{array}{l }
P\dnrm{GM}(t|e) = score(\mathbf{e})\\ \\
\mbox{Entity-level:} \\ 
~~~\mathbf{e} = \mbox{distributed embedding of} ~ e \\ \\

\mbox{Word-level:} \\
~~~\mathbf{e} = g(\mbox{words of the name of } e)\\\\

\mbox{Character-level:}\\
~~~\mathbf{e} = g(\mbox{characters of the name of } e)

\end{array} $

&  $\begin{array}{l}
\mbox{MIML-ATT:} \\
~~~P\dnrm{CM}(t|e) = score(\mathbf{a_t}) \\ 
~~~\mathbf{a_t} = \sum_i \alpha_i \mathbf{c_i} \\ \\ 
\mbox{MIML-MAX/MEAN:} \\
~~~P\dnrm{CM}(t|e) = \mbox{max}/\mbox{mean}_i P(t|c_i) \\ 
~~~P(t|c_i) = score( \mathbf{c_i})\\ 
\\
\mathbf{c_i}= g(\mbox{words of } c_i)
\end{array} $ \\
\hline
\multicolumn{2}{c}{$\begin{array}{c}
\mbox{Joint Model} \\
P(t|e) = \frac{1}{2}\big(P\dnrm{GM}(t|e) +  P\dnrm{CM}(t|e)\big)
\end{array}$
}
\end{tabular}
}
}
\caption{An overview of our global, context and joint models.
$g$ is a neural network function, e.g., a feed-forward neural network.
$e$ is an entity.
$c_i$ is the $i$-th context of $e$.
$\alpha_i$ is a scalar.
In our setup, we take the most frequent mention of $e$ in the corpus as the name of $e$.
MIML-ATT and MIML-MAX/MEAN are our different multi-instance multi-label 
models.}
\tablabel{overview}
\end{table}

\subsection{Global Model}
\seclabel{gm}

\begin{figure}[htp] 
\centering{
\includegraphics[scale=0.60]{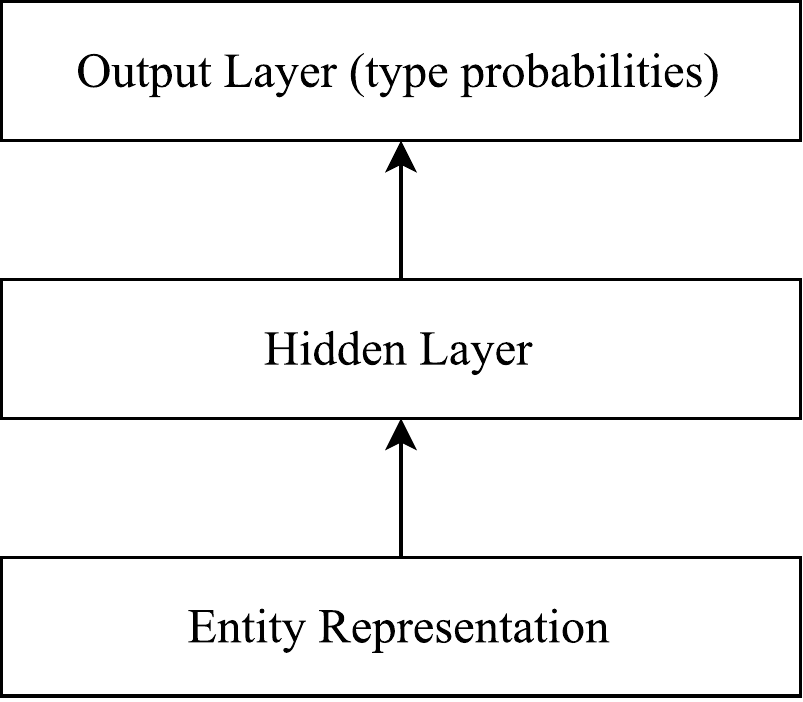}}
\caption{In the global model, a multi-layer perceptron is applied on the entity representation learned from
contexts or/and name of the entity.}
\label{fig:mlp}
\end{figure}

In the \gmodel{}, we learn $P(t|e)$ by first learning a distributed representation $\mathbf{e}$ of entity $e$.
Then, a multi-layer perceptron (MLP) with one hidden layer is applied with the output layer of size $|T|$. 
The schematic diagram of the MLP is shown in \figref{mlp}.
Unit $t$ of this layer outputs 
the probability for type $t$:
\begin{equation} 
\eqlabel{MLPscore}
P\dnrm{GM}(t|e)
= \sigma\Big(\mathbf{W}\dnrm{out}  f\big(\mathbf{W}\dnrm{in}\mathbf{e}\big)\Big)
\end{equation}
where 
$\textbf{W}\dnrm{in} \in \mathbb{R}^{h\times d} $ is the weight matrix from
$\mathbf{e} \in \mathbb{R}^d$ to the hidden layer with size $h$. 
$f$ is the rectifier function. 
$\mathbf{W}\dnrm{out} \in \mathbb{R}^{|T| \times h} $ is the weight matrix
from hidden layer to output layer of size $|T|$.
$\sigma$ is the sigmoid function: $\sigma(x) = 1/(1+e^{-x})$ that converts the value $x$ to a value in $[0,1]$.
Our objective is binary cross entropy summed over types:
\begin{equation*}
L = \sum_{t}{\mbox{BCE}(\mathbf{y}_t, \mathbf{p}_t)}
\end{equation*}
where $\mathbf{y}_t$ is the truth and
$\mathbf{p}_t$ the prediction for type $t$.

The key difficulty when computing $P(t|e)$
is in 
learning a good representation for entity $e$. 
We make use of contexts and name
of $e$
to represent its feature vector
on the three levels of entity, word and character.

\subsubsection{Entity-Level Representation}
\figlabel{elr}
Distributional representations or embeddings are commonly used
for words. 
The underlying hypothesis is that words with similar
meanings tend to occur in similar contexts \cite{harris54}
and therefore cooccur with similar context words.
We can extend the distributional hypothesis to entities (cf.,  \citeR{Wang14joint}):
entities with similar meanings tend to have similar contexts.
Thus, we can learn a $d$ dimensional embedding $\mathbf{e}$ of entity 
$e$ from a corpus in which all mentions of the entity have been replaced by
a special identifier.
We refer to 
these entity vectors as the \emph{entity-level representation} (ELR).

In previous work, order information of context words (relative position 
of words in the contexts)
 was generally ignored and objectives similar to the
SkipGram (henceforth: \emph{SKIP}) model were
used to learn $\mathbf{e}$.  However, 
the bag-of-word context is difficult to distinguish for
pairs of types that have similar context words like (``{restaurant}'',  ``{food}'') and
(``{author}'', ``{book}''). 
This suggests that \emph{using order aware
embedding models is important for entities}.
Therefore, we apply \citeS{ling15embeddings} extended
version of SKIP, Structured SKIP (SSKIP). It
incorporates the order of context
words into the objective
and outperforms SKIP in entity typing \cite{derata16,figment17}.

\subsubsection{Word-Level Representation}
\seclabel{wlr}
Words inside entity names are important sources
of information for typing entities. 
We define the word-level representation (WLR)
as the \emph{average 
of the embeddings of the words} that constitute the entity name 
$
\mathbf{e} = 1/n \sum_{i=1}^n \mathbf{w}_i
$
\noindent where
$\mathbf{w}_i$ is the embedding
of 
the $i\uprm{th}$ word of an entity name of length
$n$. 
We consider the canonical name of an entity in the KB  to compute this representation.
We opt for simple averaging since entity names
often consist of a small number of words with
clear semantics. Thus, averaging is a promising way of
combining the information that each word contributes.

The word embedding, $\mathbf{w}$,
itself can be learned from models with different granularity levels.
Embedding models that consider words as atomic units in
the corpus, e.g., SKIP and SSKIP,
are word-level.
On the other hand, embedding models 
that represent words with their character ngrams, 
e.g., \textsc{fasttext}  \cite{subword16},
are subword-level.
Based on this, we consider and evaluate \emph{word-level WLR (WWLR)}
and \emph{subword-level WLR (SWLR)} in this
paper.\footnote{Subword models have properties of
both  character-level models (subwords are character ngrams) and
of word-level models (they do not cross boundaries between
words). They probably could be put in either category, but
in our context fit the word-level category better
because we see the level of granularity with respect to the entities 
and not words.}

\subsubsection{Character-Level Representation}
\seclabel{name2type} 

For computing the \emph{character-level representation} (CLR), 
we design models that type an entity based on the sequence of
characters of its name.  Our hypothesis is that names of
entities of a specific type often have similar character
patterns.  
Entities of type ``{ethnicity}'' often
end in ``ish'' and ``ian'', e.g., ``Spanish'' and
``Russian''. Entities of type ``{medicine}'' often end in
``en'': ``Lipofen'', ``acetaminophen''. 
Also, some types tend to have specific cross-word shapes in
their entities, e.g., 
``{person}'' names usually consist of two or three words, 
or ``{music}'' names are usually long, containing several words.

The first layer of the character-level models is a
\emph{lookup table} that maps each character to an
embedding of size $d_c$. These embeddings capture
similarities between characters, e.g., similarity in the type of
phoneme encoded (consonant/vowel) or similarity in the case
(lower/upper).
The output of the lookup layer for an entity name
is a matrix $\mathbf{C} \in \mathbb{R}^{l
  \times d_c}$ where $l$ is the maximum length of a name and
all names are padded to length $l$. This length $l$ includes
special start/end characters that bracket the entity name.

We experiment with two architectures to produce
character-level representations in this paper: fully connected feed-forward (FF)
and convolutional neural networks (CNNs). 

\textbf{FF}
simply concatenates all rows
of matrix $\mathbf{C}$; thus, $\mathbf{e} \in \mathbb{R}^{d_c. l}$.

The \textbf{CNN}
uses $n$ filters of different window widths $w$
to narrowly convolve $\mathbf{C}$.
For each filter $\mathbf{H} \in \mathbb{R}^{d_c\times w}$,
the result of the convolution of $\mathbf{H}$ over matrix $\mathbf{C}$
is feature map $\mathbf{m} \in \mathbb{R}^{l-w+1}$:
\[\mathbf{m}[i] = \mbox{g}(\mathbf{C}_{[:, i : i + w - 1]} \odot \mathbf{H} + b)\]
where 
$g$ is the rectifier function ($g(x) = \max(0, x))$,
$b$ is the bias,
$\mathbf{C}_{[:, i : i + w - 1]}$ are the columns $i$ to $i + w - 1$
of $\mathbf{C}$,
 $ 1\leq w\leq k$ are the window widths we consider
 and $\odot$   is the Frobenius product. 
Max pooling then gives us one feature for each
filter. The concatenation of all these features 
is our representation: 
$\mathbf{e} \in \mathbb{R}^{n}$.
An example CNN architecture is shown in \figref{cnn}. 

\begin{figure}[h] 
\centering{
\includegraphics[scale=0.55]{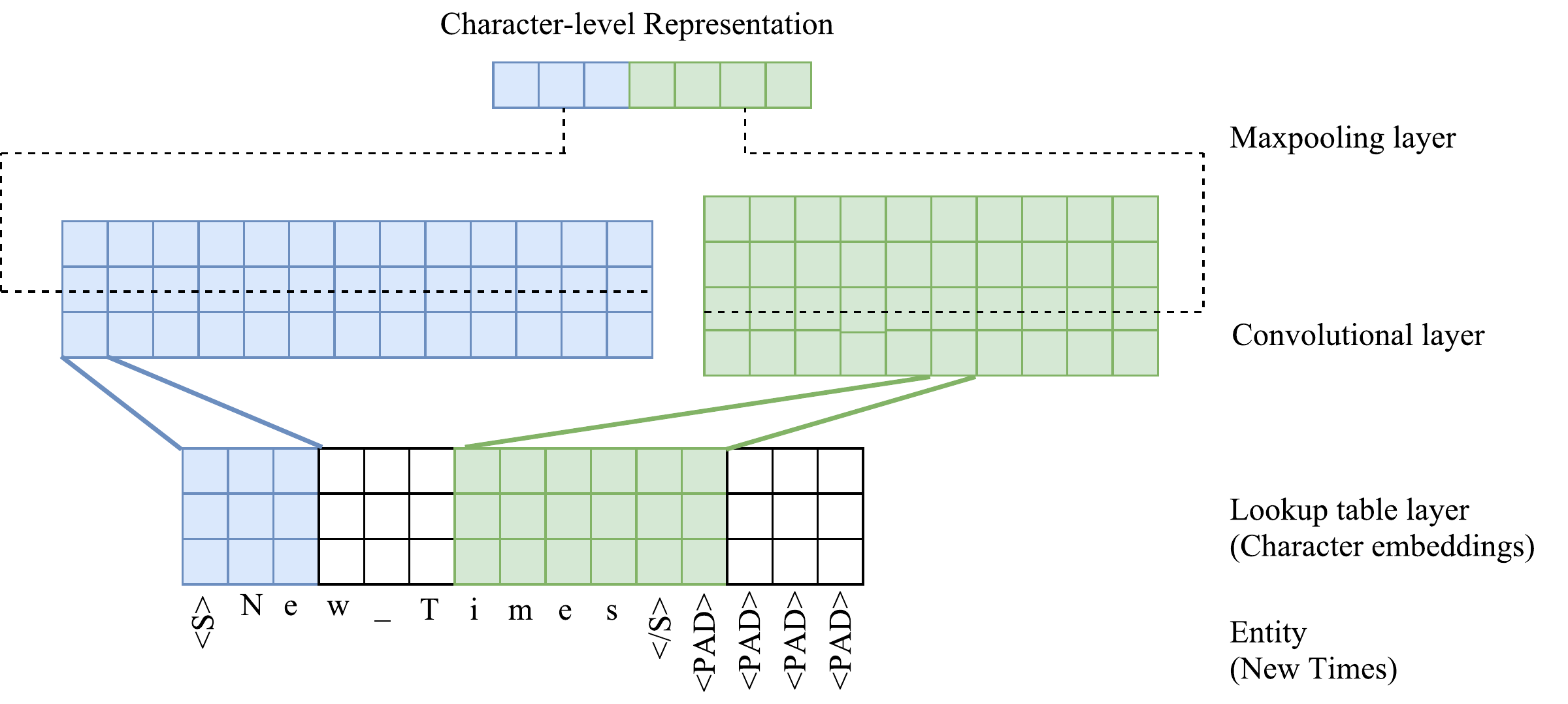}}
\caption{Example architecture for the 
character-level CNN with max pooling.
The input is ``New Times''.
Start and end symbols are appended to the input 
and it is padded by the pad symbol to a fixed size (here it is 15).
Character embedding size is three.
There are three filters of width 3 and 
four filters of width  6.
\figlabel{cnn}}
\end{figure}

\subsubsection{Multi-Level Representations}
Our different levels of representations can
provide complementary information 
about entities.

\textbf{WLR and CLR}. 
Both WLR models, SWLR and WWLR, do not have access to
the cross-word character ngrams 
of entity names while CLR models do.
Also, CLR is task specific by training
on the entity typing dataset
while WLR is generic.
On the other hand, WWLR and SWLR models 
have access to information that CLR ignores:
the tokenization of entity names into words and 
embeddings of these words. 
It is clear that words 
are
particularly important character sequences since they
often correspond to linguistic units with clearly
identifiable semantics -- which is not true for most
character sequences.
For many entities, 
their constituting words are a better basis for typing
than the 
character sequence. For example, even
if ``nectarine'' and ``compote'' did not occur in any names
in the training corpus, we can still learn good word
embeddings from their non-entity occurrences. This might
allow us to correctly type the entity ``Aunt Mary's
Nectarine Compote''
as ``{food}'' based on
the sum of the word embeddings.

\textbf{WLR/CLR and ELR}. 
Representations from entity names, i.e., WLR and CLR, by themselves are limited because 
many classes of names can be used for different types of entities;
e.g., person names do not contain hints as to whether
they are referring to a politician or athlete.
In contrast, the ELR
embedding is based on the contexts on an entity,
which are often informative  
 for each entity
and can distinguish politicians from  athletes.
On the other hand, 
not all  entities
have sufficiently many informative contexts in the corpus.
For these entities, their name can be a complementary 
source of information and character/word-level 
representations can increase typing accuracy.

\begin{figure}[h] 
\centering{
\includegraphics[scale=0.6]{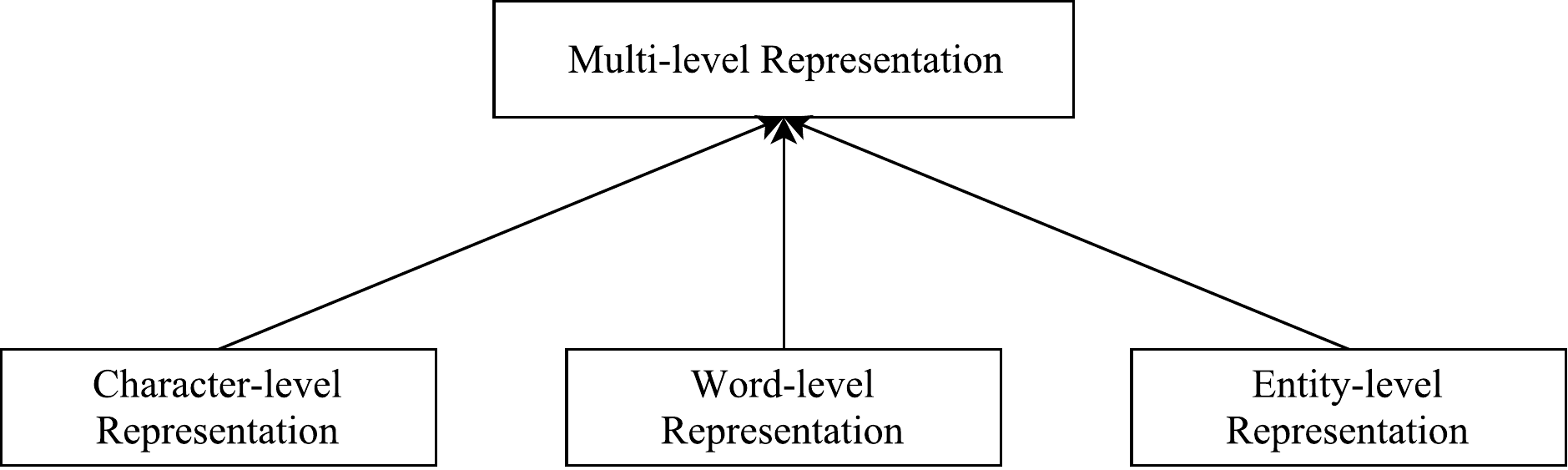}}
\caption{Multi-level representations of entities are the concatenation of a subset of 
character, word and entity-level representations.
These representations are then given to the MLP in \figref{mlp}.}
\label{fig:multilevel}
\end{figure}

Thus, we introduce \textbf{multi-level} models that use combinations of the
three levels.  Each multi-level model concatenates several
levels.  We train the constituent embeddings as follows.
WLR and ELR are computed as described above and are not
changed during training.  CLR -- produced by one of the
character-level networks described above -- is initialized
randomly and then tuned during training.  Thus, it can focus
on complementary information related to the task that is not
already present in the other levels.
The schematic diagram of our multi-level representation is shown 
in \figref{multilevel}. 

\subsection{Context Model}
\seclabel{cm}
For the  \cmodel{} (CM), we first learn a probability function
$P(t|c)$ for individual contexts $c$ in the
corpus.  $P(t|c)$ is the probability that an entity occurring in context $c$ has type
$t$.
For example, 
consider the contexts $c_1$ =
``he served SLOT cooked in wine'' 
and $c_2$ = ``she loves SLOT more than anyone''.
SLOT marks the
occurrence of an entity and it also shows that we do not
care about the entity mention itself but only its context.
For the type $t$ = ``food'',
$P(t | c_1)$ is high whereas 
$P(t | c_2)$ is low.
This example
demonstrates that some contexts of an entity like ``beef'' allow
specific inferences about its type whereas others do not.
Based on the context probability function $P(t|c)$, we then
compute the entity-level CM probability function $P(t|e)$.

More specifically, consider $B = \{c_1 ,c_2 , \dots ,c_{q}\}$ as the set of $q$
contexts of $e$ in the corpus.
Each $c_i$ is an instance of $e$ and since $e$ can have
several labels, it is a multi-instance multi-label (MIML) learning
problem.
We address MIML using neural networks 
by representing each context as a vector $\mathbf{c}_i \in \mathbb{R}^{h}$,
and learn $P(t|e)$ from the set of contexts of entity $e$.
In the following, we describe our MIML algorithms that work on the contexts representations
to compute $P(t|e)$.

\subsubsection{Distant Supervision}
The basic way to estimate $P(t|e)$ is based on distant supervision 
with learning the type probability of each $c_i$ individually,
by making the assumption that each $c_i$ expresses all 
labels of $e$.
Therefore, we define the context-level probability
function as:
\begin{equation}
P(t|c_i) = \sigma (\mathbf{w}_t^T \mathbf{c}_i + b_t)
\eqlabel{cm}
\end{equation}
where $\mathbf{w}_t \in \mathbb{R}^{h}$ is the output weight vector and 
$b_t $ is the bias scalar for type $t$.
The cost function is defined based on binary cross entropy:
\begin{equation}
L(\theta) = 
\sum_{c} \sum_{t}{\mbox{BCE}(\mathbf{y}_t, P(t|c)})
\eqlabel{loss} 
\end{equation}
where  $\mathbf{y}_t$ is 1 if entity $e$ has type $t$
and 0 otherwise. 
To compute  $P(t|e)$ at prediction time, i.e., 
$P^{\mbox{\hbox{\scriptsize pred}}}_{\mbox{\scriptsize CM}}(t|e)$,
the context-level probabilities need to be aggregated. 
Average is the usual way of doing that:
\begin{equation}\eqlabel{simpleaverage}
P^{\mbox{\hbox{\scriptsize pred}}}_{\mbox{\scriptsize CM}}(t|e) = \dfrac{1}{q} \sum_{i = 1}^{q} P(t|c_i)
\end{equation}

\subsubsection{Multi-instance Multi-label (MIML)}
The distant supervision assumption is
that \emph{all} contexts of an entity
with type $t$ are contexts of $t$; e.g.,
we label all contexts mentioning
``Barack Obama'' with all of his types.
Obviously, the labels
are incorrect or \emph{noisy} for some contexts.
Multi-instance multi-label (MIML) learning addresses this problem.
We apply MIML to fine-grained ET  for the first time.
Our assumption is: if entity $e$ has type $t$, then
there is at least one  context of $e$ in the corpus in which
$e$ occurs as type $t$.
So, we apply this assumption during training
with the following estimation of the type probability of an entity:
\begin{equation}
P\dnrm{CM}(t|e) = \max_{1 \le i \le q} P(t|c_i)
\eqlabel{miml}
\end{equation}
which means we take the \textbf{maximum} probability of type $t$
over all contexts of entity $e$ as $P(t|e)$.
We call this approach \textbf{MIML-MAX}.

%
%

\begin{figure}[h] 
\centering{
\includegraphics[scale=0.35,page=1]{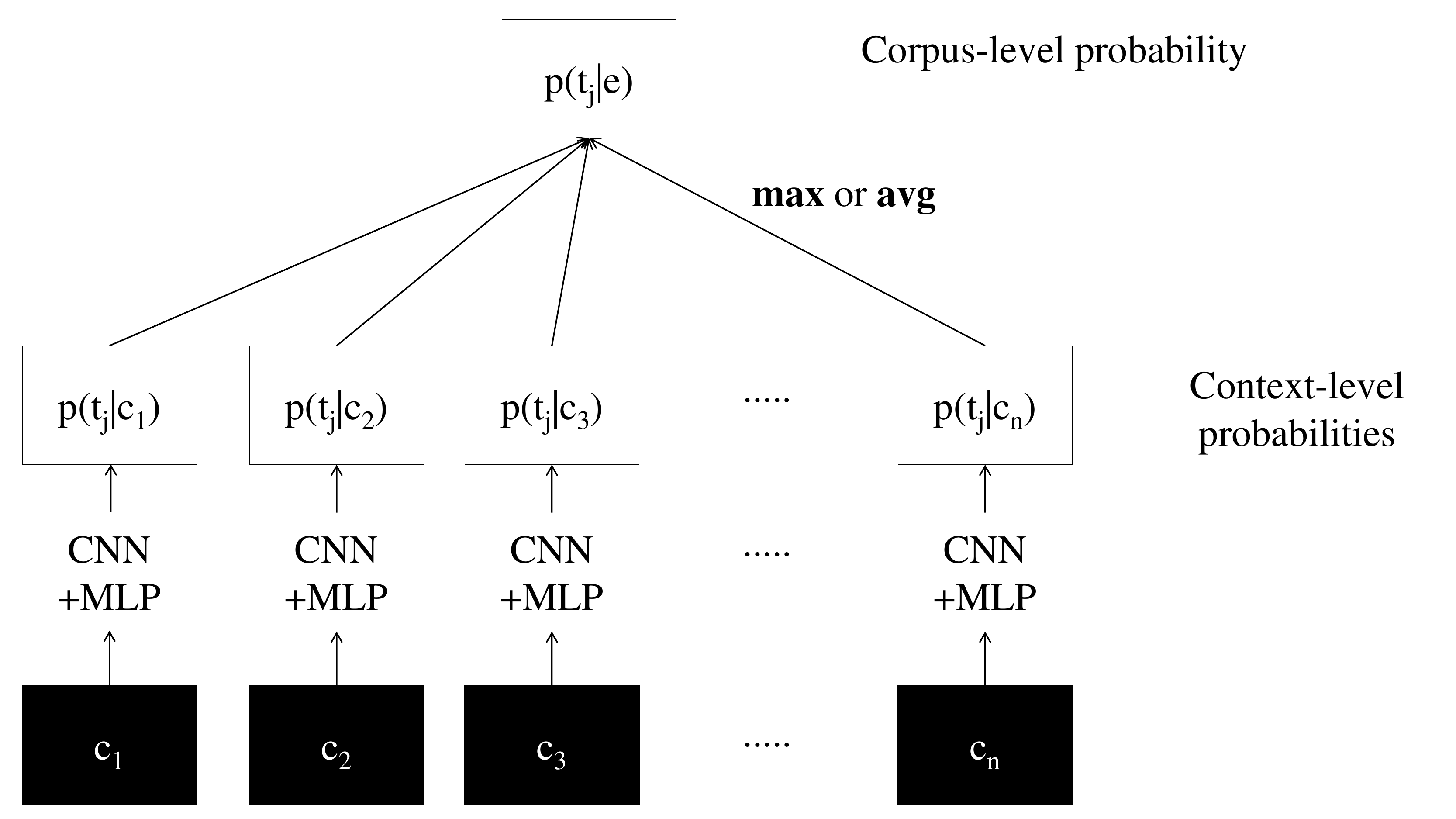}}
\caption{The context model with distant supervision, MIML-MAX, or MIML-AVG.
It uses maximum or average to compute corpus-level 
type probabilities for an entity from its context-level probabilities.
In the case of  multi-instance multi-label learning (MIML-MAX or MIML-AVG), the aggregation function (max or avg)
is applied in both training and prediction, otherwise (i.e., distant supervision) it is only applied at the prediction time. 
}
\label{fig:miml}
\end{figure}

Distant supervision makes the assumption that each
  instance is relevant. In MIML-MAX, we correct this by changing this
  assumption to ``one instance is relevant''.
This means that MIML-MAX selects the most confident context for prediction of each type.
Apart from missing information, this can be especially harmful if the entity annotations in the corpus are
the result of an entity linking system. 
In that case, 
the most confident context might be wrongly linked to the entity.
So, it can be beneficial to leverage all contexts into the
final prediction.
\textbf{Averaging} the type probabilities
of all contexts of entity $e$ is the simplest way of doing
this (cf.\ \eqref{simpleaverage}):
\begin{equation}
P\dnrm{CM}(t|e) = \dfrac{1}{q} \sum_{i = 1}^{q} P(t|c_i)
\eqlabel{avg_inst}
\end{equation}
We call this approach \textbf{MIML-AVG}.
We also propose a combination of the maximum and average,
which uses MIML-MAX (\eqref{miml}) in training 
and MIML-AVG (\eqref{avg_inst}) in prediction.
We call this approach \textbf{MIML-MAX-AVG}.
An illustration of these MIML approaches is depicted in \figref{miml}.

\begin{figure}[h] 
\centering{
\includegraphics[scale=0.4,page=2]{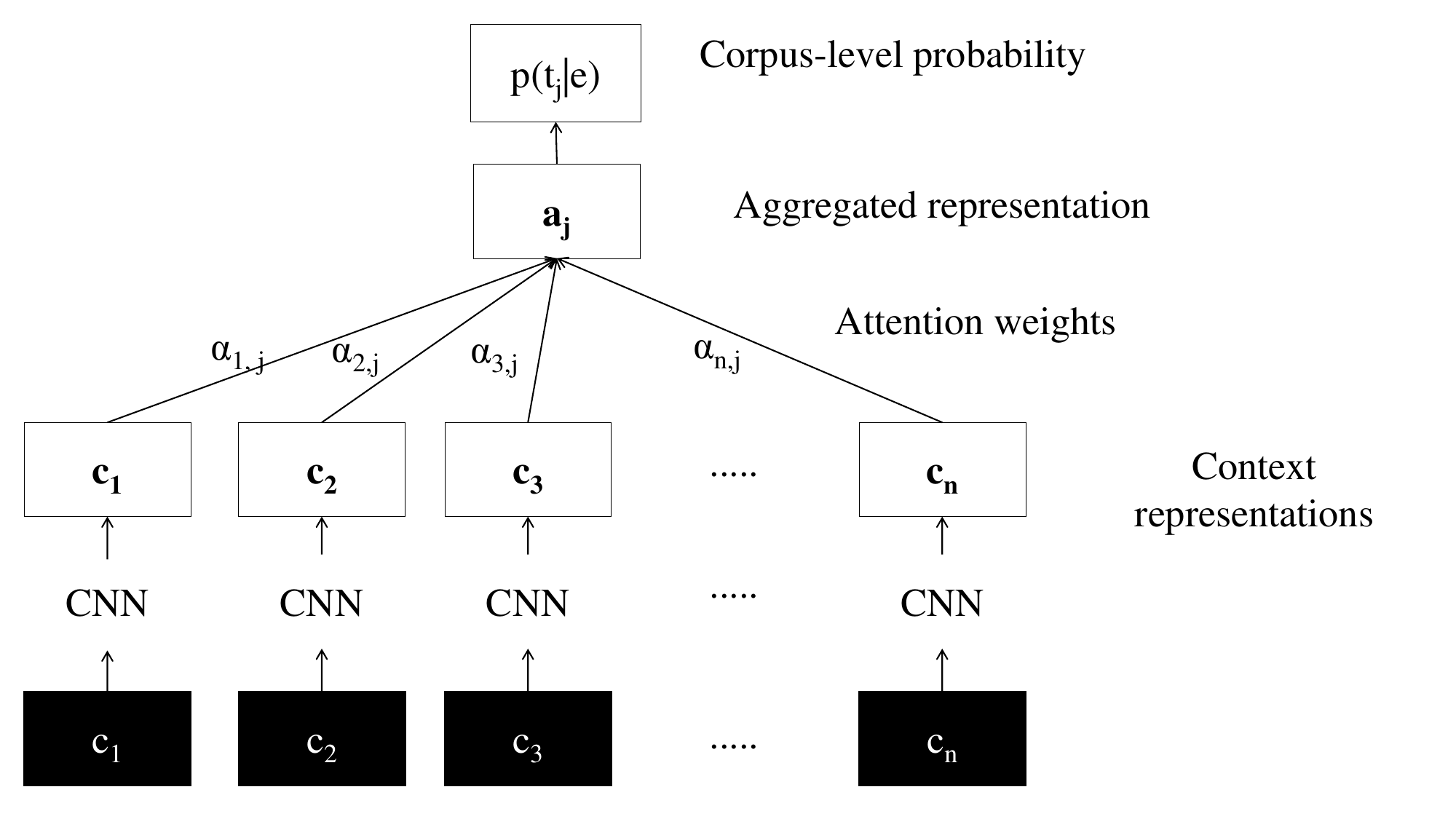}}
\caption{The context model with MIML-ATT. The corpus-level probability of a type is computed
from type specific aggregated representation of the entity, which is computed using 
attention weights of the type on each of entity contexts. 
}
\label{fig:att}
\end{figure}

MIML-AVG treats every context equally which might
be problematic since many contexts are irrelevant 
for a particular type.
A better way is to weight the contexts according to their
similarity to the types.
Therefore, we propose using selective \textbf{attention}
over contexts as follows and call this approach \textbf{MIML-ATT}.
MIML-ATT is the multi-label version of the selective attention method 
proposed by \citeA{lin2016}. 
To compute the type probability for $e$, we define:
\begin{equation}
P\dnrm{CM}(t|e) = \sigma (\mathbf{w_t}^T \mathbf{a_t} + b_t) 
\end{equation}
where $\mathbf{w_t} \in \mathbb{R}^{h}$ is the output weight vector and 
$b_t $ the bias scalar for type $t$, and 
$\mathbf{a_t}$ is the aggregated representation of all contexts $c_i$
of $e$ for type $t$, computed as follows:
\begin{equation}
\mathbf{a_t} = \sum_{i}{ \alpha_{i,t} \mathbf{c_i}}
\end{equation}
where $\alpha_{i,t}$ is the attention score of context $c_i$ for type $t$
and
$\mathbf{a_t} \in \mathbb{R}^{h}$ can be interpreted as the representation of entity $e$
for type $t$.
$\alpha_{i,t}$ is defined as:
\begin{equation}
\alpha_{i,t} = \dfrac{\exp(\mathbf{c_i} ^T\mathbf{M} \mathbf{t})}
{\sum_{j=1}^{q}{\exp(\mathbf{c_j}^T \mathbf{M} \mathbf{t}})}
\eqlabel{alpha}
\end{equation}
where $\mathbf{M} \in \mathbb{R} ^ {h \times d_t}$ is a weight matrix that 
measures the similarity of $\mathbf{c}$ and $\mathbf{t}$. 
$\mathbf{t} \in \mathbb{R}^{d_t}$ 
is the representation of type $t$.
Figure \ref{fig:att} illustrates MIML-ATT.

\tabref{mimlmodels} summarizes the differences
of our MIML methods with respect to the aggregation function they use
to get corpus-level probabilities.
For optimization of all MIML methods, we use the binary cross entropy loss function,  
\begin{equation}
L(\theta) = \sum_{e} \sum_{t} \mbox{BCE}(\mathbf{y}_t, P(t|e))
\eqlabel{miml_loss} 
\end{equation}
In contrast to the loss function of distant supervision in \eqref{loss}, 
which iterates over all \emph{contexts},
we iterate over all \emph{entities} here.

\begin{table}[h]
\begin{center}
\footnotesize
\begin{tabular}{l|c|c}
Model & Train & Prediction \\ 
\hline 
MIML-MAX & MAX & MAX \\ 
\hline 
MIML-AVG & AVG & AVG \\ 
\hline 
MIML-MAX-AVG & MAX & AVG \\ 
\hline 
MIML-ATT & ATT & ATT
\end{tabular} 
\caption{Different MIML algorithms for entity typing, and the aggregation
function they use to get corpus-level  probabilities.}
\tablabel{mimlmodels}
\end{center}
\end{table}

\subsubsection{Context Representation}
\seclabel{model}
To produce a high-quality
context representation $\mathbf{c}$, 
we use two neural network architectures, fully connected feed-forward (FF) and convolutional neural networks (CNNs).
The first layer of both architectures is a
\emph{lookup table} that maps each word in $c$ to an 
embedding of size $d$. 
If there is another training entity in the context, we replace it 
with its notable type to get better generalization. 
The output of the lookup layer 
is a matrix $\mathbf{E} \in \mathbb{R}^{d
  \times s}$ (the embedding layer), 
  where $s$ is the context size (a fixed number of words).
\textbf{FF} then simply concatenates all rows
of matrix $\mathbf{E}$; thus, $\phi(c)\in \mathbb{R}^{d*s}$.

\textbf{CNN}
uses $n$ filters of different window widths $w$
to narrowly convolve $\mathbf{E}$.
For each of the $n$ filters $\mathbf{H} \in \mathbb{R}^{d\times w}$,
the result of applying $\mathbf{H}$ to  matrix $\mathbf{E}$
is a feature map $\mbox{m} \in \mathbb{R}^{s-w+1}$:
\begin{equation}
\mathbf{m}[i] = g(\mathbf{E}_{:, i : i + w - 1} \odot \mathbf{H})
\end{equation}
where 
$g$ is the rectifier function,
$\odot$ is the Frobenius product,
$\mathbf{E}_{:, i : i + w - 1}$ are the columns $i$ to $i + w - 1$
of $\mathbf{E}$ and
 $ 1\leq w\leq k$ are the window widths we consider.
Max pooling then gives us one feature for each
filter and the concatenation of those features 
is the CNN representation of $c$.
We apply the CNN to the left and right context of the entity mention
and the concatenation $\phi(c) \in \mathbb{R}^{2n}$ is the output of the CNN layer. 

$\phi(c)$ is then fed  into a multi-layer perceptron (MLP) for both architectures
(FF and CNN)
to get the final context representation $\mathbf{c} \in \mathbb{R}^{h}$:
\begin{equation}
\mathbf{c} = \tanh\Big(\mathbf{W}_h \phi(c)\Big)
\end{equation}

\subsection{Joint Model}
\seclabel{joint}
Global model
and \cmodel{} have complementary strengths and weaknesses.

The strength of CM is that it is a direct model of the contexts
 in which
the entity occurs. This is also the way a human would
ordinarily do entity typing for an unknown entity: they would determine if a
specific context in which the entity occurs implies that the
entity is, say, an author or a musician and type it
accordingly. The order of words  is of critical importance
for the accurate assessment of a context and CM takes it
into account. A well-trained CM will also work for cases
for which GM is not applicable. In particular, if the KB
contains only a small number of entities of a particular
type, but the corpus contains a large number of contexts of
these entities, then CM is more likely to generalize
well.

One notable weakness of CM is that a large proportion of
contexts does not contain sufficient information to infer
all types of the entity. For example, based on our distantly
supervised training data, we label every context of
``Obama'' with ``{politician}'' and ``{author}'' and Obama's
other types in the KB.
Multi-instance learning algorithms mitigate this weakness to some degree, but 
at the extra cost of more model parameters.

The main strength of GM is that it 
bases its decisions on the entire evidence available in the
corpus. This makes it more robust. 
Also, GM models the problem as a supervised task and its dataset consists 
of labeled entities with their types.
This makes it more efficient compared to the distantly supervised 
approach used in CM.

The disadvantage  of GM
is that it is less likely to work well for non-dominant
types of an entity that might be swamped by dominant types.
For example, the author contexts of ``Obama'' may be swamped by the
politician contexts and the overall context signature of the
entity ``Obama'' may not contain enough signal to infer that
he is an author.
Also for an unknown entity, 
the embedding learning model needs to be retrained to get the 
entity-level representation, which makes the GM application less efficient for new or emerging entities.

Since GM and CM models are complementary, we expect a combined model to work better. We test this hypothesis for the
simplest possible joint model (JM), which averages the type probabilities of the
two individual models for each entity as:
\begin{equation}
P\dnrm{JM}(t|e) = \frac{P\dnrm{GM}(t|e) + P\dnrm{CM}(t|e)}{2} 
\end{equation}

\section{Experimental Setup and Results}
\seclabel{exp}
In this section, we first describe the different setups we use to do our experiments.
Next, we present the results of our models followed by further analysis. 

\subsection{Setup}
In this section, we explain the datasets, evaluation metrics and other setups we follow to do our
experiments. We also describe the baselines, which we compare our models against. 

\subsubsection{Corpus}
We select a subset of about 7.5
million web pages, taken from the first segment of
ClueWeb12 \cite{clueweb12url},
from different crawl types: 1 million Twitter links, 120,000 WikiTravel pages and 6.5 million web pages.
This corpus is preprocessed by eliminating HTML tags,
replacing all numbers with ``7'' and all web links and email
addresses with ``HTTP'', filtering out sentences with length
less than 40 characters, and finally doing a simple tokenization.  
We  merge the text with the
FACC1 annotations.
The resulting
corpus has  4 billion tokens and  950,000
distinct entities. We use
the 2014-03-09 
Freebase data dump 
as our  KB.

\subsubsection{Entity Datasets}
We consider all entities in the corpus  whose notable types can be mapped 
to one of the 113 FIGER types, based on the mapping provided by FIGER. 
750,000 such entities form our set of entities.
10 out of 113 FIGER types have no entities in this
set.\footnote{The reason is that
the FIGER mapping uses Freebase user-created
classes. The 10 missing types are not
the notable type of any entity in Freebase.}
We then select a subset of 200,000 entities in a process which is explained 
by \citeA{figment15}.
We split the entities   into train (50\%), dev (20\%) and test (30\%) sets.  
The average and median number of FIGER types of the training
entities  are 1.8 and 2, respectively.\footnote{The entity datasets are available at
\url{http://cistern.cis.lmu.de/figment}.
}

\subsubsection{Context Sampling} 
For the \cmodel{},
we create train', dev' and
test' sets of \emph{contexts} that correspond to 
train, dev and test sets of \emph{entities}. 
Because the number of contexts is unbalanced for both
entities and types and also to
accelerate training and testing, we downsample contexts.
For the set train', we use the notable type feature of Freebase:
For each type $t$, we take contexts from the mentions of $t$.

Next, if the number of contexts for $t$ is larger than a minimum, 
we sample the contexts based on the number of training entities of $t$. 
We set the minimum to 10,000 and constrain the number of samples for each $t$ to 20,000. 
Also, to reduce the effect of distant supervision, entities with fewer distinct types
are preferred in sampling to provide discriminative contexts for their notable types.
For test' and dev' sets, we sample 300 and 200 random contexts, respectively, for each entity.

\subsubsection{Baselines}
We extend our previous work in corpus-level fine-grained entity typing.
JOINT-BASE1, JOINT-BASE2 and ELR+SWLR+CLR(CNN) correspond to the best models by  \citeA{figment15}, \citeA{noise17} and  \citeA{figment17}, respectively.
We also add some hand-crafted feature-based baselines.

We implement the following two feature sets from the
literature as a \emph{hand-crafted baseline} for
our character-level and word-level 
GM models.  (i) \emph{BOW}: individual words of entity name (both
as-is and lowercased); (ii) \emph{NSL} (ngram-shape-length): shape and length of the
entity name (cf., \citeR{ling2012fine}), character
$n$-grams, $1 \leq n \leq n\dnrm{max}, n\dnrm{max}=5$ (we
also tried $n\dnrm{max}=7$, but results were worse on dev) and
normalized character $n$-grams: lowercased, digits
replaced by ``7'', punctuation replaced by ``.''.
These features are represented as a sparse binary vector 
$\mathbf{e}$ that forms the input to the architecture in
\figref{mlp}.

The other baseline is using an existing mention-level entity typing
system, \emph{FIGER} \cite{ling2012fine}.
FIGER uses a wide variety of  features
on different levels (including
parsing-based features)
from contexts of entity mentions
as well as the mentions themselves and returns a score for each
mention-type instance in the corpus.
We provide the ClueWeb/FACC1 segmentation of entities,
so FIGER does not need to recognize entities.\footnote{Mention typing is separated from recognition in FIGER model. So it can use our segmentation of entities.}
We use the trained model provided by the authors
and normalize FIGER scores using
softmax to make them comparable for aggregation.
We experimented with different aggregation functions
(including maximum and $k$-largest scores for a type).
The average of scores gave us the best result on dev.
 We call this baseline AGG-FIGER:
aggregated version of FIGER \cite{ling2012fine}.

\subsubsection{Evaluation Metrics}
Evaluation of multi-label classification is more complicated than the common single-label 
setting because each example can belong to several labels at the same time. 
To address this, we use two types of metrics for evaluation:
example-based  and label-based. 
In the example-based measures, an average value is computed based on the 
evaluation values of each test example.
In the label-based measures, the performance of a classifier is evaluated for each
label and then averaged over all labels. 
Each measure can further be categorized into ranking or classification. 
In the ranking measures, the evaluation shows how well the models rank 
labels for examples (for the example-based measures), or, 
rank examples for labels (for the label-based measures). 

Since our problem is multi-label classification, we adapt some of the 
measures from the literature. In our setting, entities are examples and
types are the labels.
The classification metrics measure the quality of the thresholded assignment decisions
produced by the models. 
These measures more directly express how well
\textsc{figment} would succeed in enhancing the KB with new
information since for each pair $(e,t)$, we have to make a
binary decision about whether to put it in the KB or not.
The decisions are then compared to the gold KB information, with the assumption that
KB information is incomplete. 
The assignment decision is made based on thresholds, one
per type, for each $P(t|e)$. 
We select the threshold that maximizes 
$F_1$ of entities assigned to the type on dev.
In the following, we define the metrics.

\emph{Example-based metrics:}
Two ranking and three classification measures are considered for the example-based
evaluation.
The ranking measures are (i) precision at 1 (P@1): percentage of entities 
whose top-ranked type is correct; (ii) breakeven point (BEP, \citeR{boldrin2008against}): $F_1$ at the point in the ranked list at
which precision and recall have the same value.
The classification measures are 
(i) accuracy: an entity is correct if
all its types and no incorrect types are assigned to it;
(ii) micro average: $F_1$ of all type-entity assignment decisions;
(iii) entity macro average $F_1$: $F_1$ of types assigned to
an entity, averaged over entities.

\emph{Label-based metrics:}
The ranking measures are 
(i) mean average precision (MAP): mean of the averaged precision of each type; 
(ii) average precision at k (P@k): average 
of each type's precision in the top k ranked entities. 
The classification measure is:
(i) type macro average $F_1$: $F_1$ of entities assigned to
a type, averaged over types.

\subsubsection{Distributional Embeddings}
For WWLR and ELR, 
we use the Structured SkipGram (SSKIP) model in \textsc{wang2vec}
\cite{ling15embeddings} to learn 
the embeddings for words, entities and types.
To obtain embeddings for all three
in the same space, we process ClueWeb/FACC1  as follows.
For each sentence $s$,
we add three copies: $s$ itself, a copy of $s$ in which each
entity is replaced with its Freebase identifier (MID) and a
copy in which each entity (not test entities though) is replaced with an ID indicating
its notable type.
The resulting corpus contains around
4 billion tokens and 1.5 million types.

We run  SSKIP  with the setup
(100 dimensions, 10 negative samples, window size 5, and word frequency threshold of 100)\footnote{The threshold does not apply for MIDs.}
on this corpus to learn embeddings for words,
entities and FIGER types. 
For SWLR, 
we use \textsc{fasttext} \cite{subword16}
to learn word embeddings from the ClueWeb/FACC1 corpus.
We use similar settings as our WWLR SKIP and SSKIP embeddings and
keep the defaults of other hyperparameters.
Since the trained model of \textsc{fasttext} is applicable for 
new words, we apply the model to get 
embeddings for the filtered rare words as well. 

\subsubsection{Hyperparameter Values} 
Our hyperparameter values are optimized on dev.
We use AdaGrad \cite{adagrad} and minibatch training.
For each experiment, we select the best model on dev.
The values of our hyperparameters are shown in \tabref{hyp}
in the appendix. 

\subsection{Results}
We evaluate our different models using the mentioned measures. 
For the \gmodel{}, we explore different entity representations including multi-level ones as described in \secref{gm}. 
For the \cmodel{}, we analyze the performance of the baseline AGG-FIGER 
and different models including the mentioned MIML methods on 
two architectures (FF and CNN) as described in \secref{cm}.
For the joint model, we show the combination of our best global and context
models, as well as the combination of our baseline models.

\begin{table}[t]
\begin{center}
\setlength{\tabcolsep}{2pt}
{
\footnotesize
\begin{tabular}{l rl|cc|ccc|cc|ccc|cc|ccc}
&&& \multicolumn{5}{|c|}{all entities} &
  \multicolumn{5}{|c|}{head entities} &
  \multicolumn{5}{|c}{tail entities}\\
&&& P@1 & BEP & acc & mic &  mac
&  P@1 & BEP & acc & mic &  mac
&  P@1 & BEP & acc & mic &  mac \\
\hline\hline %
\multirow{12}{*}{\rotatebox{90}{\small global model (GM)}}
& 1 & MFT
         & .101 & .406 & .000 & .036 & .036
         & .111 & .406 & .000 & .040 & .039
         & .097 & .394 & .000 & .032 & .032 \\

& 2 & CLR(NSL)
         & \textbf{.643} & \textbf{.690} & .167 & .447 & .438
         & .615 & .665 & .154 & .438 & .425
         & .633 & .696 & \textbf{.180} & .439 & \textbf{.437} \\

& 3 & CLR(FF)
         & .552 & .618 & .032 & .345 & .313
         & .530 & .592 & .032 & .351 & .313
         & .543 & .621 & .031 & .330 & .304 \\

& 4 & CLR(CNN)
         & .628 & .679 & \textbf{.170} & \textbf{.471} & \textbf{.431}
         & \textbf{.598} & \textbf{.649} & \textbf{.166} & \textbf{.474} & \textbf{.423}
         & \textbf{.618} & \textbf{.685} & \textbf{.180} & \textbf{.458} & .429\\
\cdashline{2-18}
& 5 & BOW
                 & .560 & .636 & .080 & .311 & .370
                 & .512 & .591 & .075 & .301 & .354
                 & .566 & .654 & .088 & .318 & .376 \\
& 6 & WWLR
         & .691 & .727 & .168 & .553 & .500
         & \textbf{.767} & .785 & .246 & .634 & .610
         & .635 & .684 & .125 & .497 & .444 \\
& 7 & SWLR
         & .702 & .733 & .178 & .552 & .525
         & .761 & .782 & .238 & .625 & .599
         & .651 & .696 & .150 & .509 & .488 \\

& 8 & SWLR+CLR(CNN)
                & \textbf{.710} & \textbf{.754} & \textbf{.248} & \textbf{.578} & \textbf{.539}
                & .754 & \textbf{.795} & \textbf{.312} & \textbf{.658} & \textbf{.618}
                & \textbf{.669} & \textbf{.729} & \textbf{.230 }& \textbf{.529} & \textbf{.499} \\
\cdashline{2-18}
& 9 & ELR
         & .851 & .890 & .494 & .781 & .740
         & .901 & .922 & .541 & .831 & .802
         & .732 & .802 & .381 & .648 & .581 \\
& 10 & ELR+CLR(CNN)
        & .873 & .905 & .538 & .804 & .766
        & .906 & .928 & \textbf{.569} & .840 & .812
        & .784 & .841 & .453 & .713 & .648 \\
& 11 & ELR+SWLR
         & .877 & .907 & .532 & .804 & .769
         & .906 & .856 & .556 & .836 & .810
         & .802 & .856 & .466 & .727 & .668 \\

& 12 & ELR+SWLR+CLR(CNN)
         &\textbf{ .881} & \textbf{.911 }& \textbf{.548 }& \textbf{.812 }& \textbf{.776}
         & \textbf{.907} & \textbf{.929} & .567 & \textbf{.841} & \textbf{.813}
         & \textbf{.812} & \textbf{.862} & \textbf{.484} & \textbf{.738} & \textbf{.678} \\
\hline \hline
\multirow{11}{*}{\rotatebox{90}{\small context model (CM)}}
& 13&FF
         & .753 & .791 & .341 & .697 & .665
         & .781 & .806 & .415 & .757 & .722
         & .639 & .705 & .151 & .529 & .493 \\
& 14&FF+MIML-MAX
                & .757 & .806 & .230 & .602 & .562
                & .777 & .809 & .063 & .518 & .537
                & .660 & .738 & .210 & .479 & .345\\
& 15&FF+MIML-MAX-AVG
            & .763 & .807 & .369 & .721 & .686
                & .774 & .803 & .442 & .774 & .745
                & .668 & .741 & .189 & .568 & .519 \\
& 16&FF+MIML-AVG
                & .779 & .826 & .369 & .714 & .685
                & .794 & .830 & .425 & .760 & .731
                & .674 & .749 & .204 & .566 & .531\\
& 17&FF+MIML-ATT
                & \textbf{.820} & \textbf{.855} & \textbf{.403} & \textbf{.730} & \textbf{.701}
                & \textbf{.874} & \textbf{.889} & \textbf{.454} & \textbf{.781} & \textbf{.767}
                & \textbf{.681} & \textbf{.757} & \textbf{.283} & \textbf{.596} & \textbf{.553}\\
\cdashline{2-18}
& 18&CNN
                 & .784 & .820 & .394 & .722 & .693
                 & .818 & .840 & .461 & .773 & .747
                 & .657 & .726 & .210 & .563 & .523 \\
& 19&CNN+MIML-MAX
                 & .786 & .825 & .262 & .622 & .584
                 & .811 & .831 & .084 & .535 & .561
                 & .678 & .752 & .227 & .497 & .357 \\
& 20&CNN+MIML-MAX-AVG
                 & .799 & .834 & .417 & .743 & .708
                 & .818 & .839 & .484 & .792 & .839
                 & .687 & .757 & .260 & .598 & .541 \\
& 21&CNN+MIML-AVG
                 & .808 & .847 & .418 & .735 & .711
                 & .829 & .856 & .472 & .777 & .757
                 & .693 & .763 & .257 & .592 & .558 \\
& 22&CNN+MIML-ATT
                 & \textbf{.837} & \textbf{.869} & \textbf{.460} & \textbf{.753} & \textbf{.730}
            & \textbf{.894} & \textbf{.903} & \textbf{.504} & \textbf{.796} & \textbf{.792}
                & \textbf{.699} & \textbf{.771} & \textbf{.335} & \textbf{.626} & \textbf{.584} \\
\cdashline{2-18}
&23 & AGG-FIGER
                & \textbf{.811 }& \textbf{.847} & \textbf{.440 }& \textbf{.740 }& \textbf{.686}
                & \textbf{.843} & \textbf{.882} & \textbf{.530 }& \textbf{.815} & \textbf{.763}
                & \textbf{.738} & \textbf{.784 }& \textbf{.322 }& \textbf{.627 }& \textbf{.579 }\\

\hline \hline

& 24 & JOINT-BASE1
                & .857 & .889 & .507 & .789 & .746
                & .902 & .920 & .558 & .836 & .807
                & .735 & .80 & .380 & .662 & .591 \\
& 25 & JOINT-BASE2
                & .876 & .904 & .534 & .803 & .769
                & \textbf{.923} &\textbf{ .937} & .585 & \textbf{ .848} & \textbf{.830}
                & .756 & .820 & .415 & .685 & .622 \\
& 26 & JOINT
                & \textbf{.885} & \textbf{.911} & \textbf{.563} & \textbf{.819} & \textbf{.785}
                & .912 & .929 & \textbf{.586} & \textbf{.848} & .820
                & \textbf{.812} & \textbf{.858} & \textbf{.487} &\textbf{ .747} &\textbf{ .694}
\end{tabular}
}
\end{center}
\caption{
Example-based measures:
P@1, BEP,
acc (accuracy), mic (micro $F_1$) and mac (macro $F_1$) on test for all, head and tail entities.
Largest numbers in each column for each section separated with dotted lines are in bold font.}
\tablabel{example}
\end{table}

Results for the \textbf{example-based measures} 
on the test entities
for all  (about 60,000 entities), 
head (frequency $>$
100; about 12,200) and tail (frequency $<$ 5; about 10,000)
are shown in \tabref{example}.
Each row represents one of the models. 
If not mentioned explicitly, the micro $F_1$ (Micro in \tabref{example})
is the measure we talk about when comparing models.

Line 1 is the most frequent baseline, which assigns ``{person}'' to 
each entity. Lines 2-12 are different \textbf{\gmodel{}s} (GMs).
For the character-level representation of entities, CNN (line 4) works clearly better than FF (line 3).
This was expected as CNNs are good architectures to find
position-independent local features for classification.
NSL, the ngram baseline, works better than CNN in P@1 and BEP for all entities and macro $F_1$ for the tail entities.
But for all other measures, CNN is better than NSL. CNN is learning the features automatically and this is another benefit of CNN over NSL.
In word-level models (lines 5-8), SWLR (line 7) is better than WWLR  (line 6) on the tail entities and worse on the head entities. 
Both are better than the BOW baseline, because BOW cannot deal well with sparseness.
SWLR and WWLR are better than all CLR models (lines 2-4) because of their access to the word embeddings trained on
the whole corpus.
SWLR (line 7) has access to the subword information, and therefore has an embedding for each
word, resulting in better embeddings for the tail entities.
Since SWLR is better than WWLR for acc, we pick SWLR as our best WLR model.
SWLR+CLR(CNN) (line 8) improves SWLR (line 7) by around 2\%, 
implying that character-level representation add complementary information to
the word-level models.

The entity-level representation, ELR (line 9), is the most important source of information for entities. 
\figref{tsne} provides a t-SNE \cite{van08tsne} visualization of ELR embeddings of a subset of the 
entities in our dataset. Different colors denote different entity types.
The figure shows that entities of the same type are clustered together.
The results confirm this: ELR is clearly better than character-level and word-level representations (line 9 $>$ lines 1-8).
Adding CLR or SWLR to ELR (lines 10-11), improves ELR (line 9), especially for tail entities (around 7\% in micro $F_1$).
This demonstrates that for rare entities,
contextual information is often not sufficient for an
informative representation; hence, name features are helpful.
Combination of all these three levels (line 12) is the best GM.
This confirms our intuition that the levels are complementary.

\begin{figure}[h]
\centering
\includegraphics[scale=0.15]{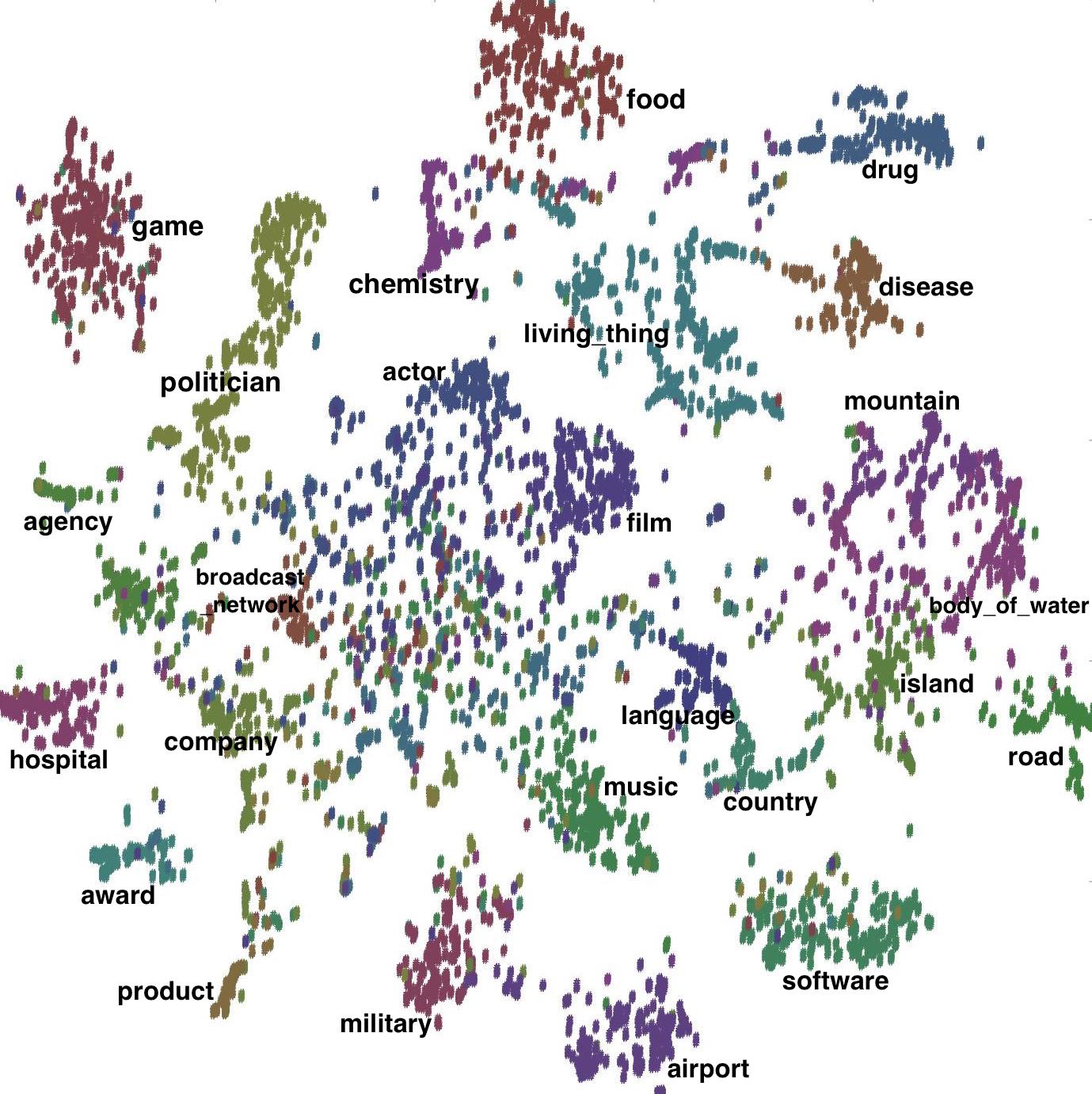}
\caption{t-SNE result of entity-level representations.} 
\figlabel{tsne}
\end{figure}


The results of \textbf{\cmodel{}s} (CMs) are presented in lines 13-23.
CNN (line 18) is better than FF (line 13);
this result for context models mirrors the result we
earlier obtained for global models (line 4 vs.\ line 3).
Lines 14-17 and 19-22 show the results of different MIML algorithms for
FF and CNN, respectively.
The order of MIML methods is consistent in both FF and CNN:
ATT $>$ MAX-AVG $>$ AVG $>$ MAX.

MAX (lines 14 and 19) is worse than its basic distantly supervised model version (lines 13 and 18).
MAX predictions are based on only one context of each entity (for each type),
and the results suggest that this is harmful for entity typing.
This is in contradiction with the previous results in RE (cf., \citeR{Zeng15emnlp}) and suggests that 
there is a difference between corpus evidence about the 
types of entities on the one hand and about relations between
entities on the other.
Related to this, MAX-AVG (lines 15 and 20) which averages the type probabilities at prediction time 
improves MAX (lines 14 and 19) by a large margin.
Averaging the context probabilities seems to be a way
to smooth the entity type probabilities. 
MAX-AVG (lines 15 and 20) models are also better than the corresponding models with AVG (lines 16 and 21)
which train and predict with averaging.
This is due to the fact that AVG gives equal weights 
to all context probabilities  in both training and prediction.
ATT (lines 17 and 22) uses weighted
contexts in both training and prediction
and that is probably the reason for its effectiveness
over all other MIML algorithms.
Overall, using attention (ATT) consistently improves 
the results of both FF and CNN models.

In line 23, we show the results of AGG-FIGER.
Compared to the neural network \cmodel{}s (lines 13-22), it is  worse than
CNN+MIML-ATT (line 22) in general, but better on head entities. 
AGG-FIGER (line 23) has access to the features extracted from the 
mentions of entities as well. This gives it more information 
than our \cmodel{}s have which only use 
the contexts of entities.
AGG-FIGER is trained using distant supervision and
its results could be improved by using our MIML methods.
We leave this for future work.
Nevertheless, all GM models with ELR (lines 9-12) are clearly better than CM models (lines 13-23)  and this shows the effectiveness of entity-level embeddings for corpus-level entity typing.

Finally, lines 24-26 show the results for three joint models. 
JOINT-BASE1  (line 24) is the joint model of the GM model in line 9 (ELR) and the CM model in line 13 (FF) and, as we already mentioned, JOINT-BASE1 corresponds to our best model in \citeA{figment15}.
JOINT-BASE2 (line 25) is the joint model of the GM model in line 9 (ELR) and the CM model in line 22 (CNN+MIML-ATT). JOINT-BASE2 corresponds to our
best model in \citeA{noise17}.
JOINT (line 26) is joining  the best GM (line 12) and the best CM (line 22) and 
is the best in all of the measures, on the whole dataset, across all of the models.
For the head entities, JOINT-BASE2 is slightly better. 
All joint models are better than their single counterparts.
This confirms our intuition that GM and CM have complementary 
information.
Compared to JOINT-BASE1,  JOINT-BASE2  improves 
CM by using CNN instead of FF and applying MIML.
Compared to JOINT-BASE2, JOINT improves GM by adding word-level
and character-level representations to the entity-level
embeddings.
The 
JOINT model is our \textsc{figment} system.  

\begin{table}[t]
\begin{center}
{
\footnotesize
\begin{tabular}{l rl|c|cc|cc|cc}
&&&& \multicolumn{2}{|c|}{all types} &
  \multicolumn{2}{|c|}{head types} &
  \multicolumn{2}{|c}{tail types}\\
& &                              & MAP
                    & Macro $F_1$ & P@50
                                 & Macro $F_1$ & P@50
                                 & Macro $F_1$ & P@50 \\
\hline
\multirow{9}{*}{\rotatebox{90}{\small global model}}
&1& MFT          & .018 & .032 & .016 & .172 & .084 & .002 & .001\\
\cdashline{2-10}
&2& CLR(NSL)   & \textbf{.277} & .320 & .506 & \textbf{.448} & .771 & .203 & .241 \\
&3& CLR(FF)       & .153 & .194  & .515 & .374 & .639 & .115 & .131 \\
&4& CLR(CNN)      & .274 & \textbf{.325}  & \textbf{.515}
& .439 & \textbf{.784} &\textbf{ .252} & \textbf{.276} \\
\cdashline{2-10}
&5& BOW          & .246 & .315 & .504 & \textbf{.427} & .759 & .233 & .272 \\
&6& WWLR          & .356 & .395  & .619  & .520 & .916 & .308 & .367\\
&7& SWLR                      & .374 & .412  & .627 & .527 & \textbf{.925} & \textbf{.339} & \textbf{.384}\\
&8& SWLR+CLR(CNN)         & \textbf{.392} & \textbf{.420 } & \textbf{.640}
& \textbf{.530} & .908 & .322 & .375\\
\cdashline{2-10}
&9&ELR                       & .631 & .599  & .788 & .773 & .988 & .465 & .527\\
&10&ELR+SWLR          & .671 & .632  & .813 & \textbf{.792} & \textbf{.993 }& .500 & .579\\
&11&ELR+CLR(CNN)  & .679 & .646 & \textbf{.824}& .789 & .992 & .\textbf{524} & \textbf{.597} \\
&12&ELR+SWLR+CLR(CNN) & \textbf{.685} & \textbf{.650}  & .820
& .798 & .989 & .523 & .593\\

\hline\hline
\multirow{11}{*}{\rotatebox{90}{\small context model}}
&13&FF    & .434 & .468 & .561 & .688 & .829 & .332 & .362 \\
&14&FF+MIML-MAX
                          & .432 & .422 & .667  & .602 & \textbf{.944} & .314 & .398\\
&15&FF+MIML-MAX-AVG
                          & .489  & .510 & .641 & .704 & .883 & .363 & .422 \\
&16&FF+MIML-AVG
                          & .470 & .499 & .595 & .700 & .853 & .349 & .400\\
&17&FF+MIML-ATT
                          & \textbf{.514} & \textbf{.530} & \textbf{.678} & \textbf{.712} & .915 & \textbf{.40}7 & \textbf{.452} \\
\cdashline{2-10}
&18&CNN & .478 & .507 & .603 & . 709 & .856 & .367 & .417\\
&19&CNN+MIML-MAX
                      & .466 & .450 & .697 & .618 & \textbf{.967 }& .358 & .454 \\
&20&CNN+MIML-MAX-AVG
                          & .542 & .552 & .691 & .725 & .895 & .425 & .494\\
&21&CNN+MIML-AVG
                          & .509 & .534 & .625 & .723 & .843 & .394 & .453\\
&22&CNN+MIML-ATT
                      & \textbf{.561} & \textbf{.561} & \textbf{.723} & \textbf{.736} & .951 & .\textbf{436} & \textbf{.496} \\
\cdashline{2-10}
&23&AGG-FIGER
                 & .59 & .592 & .756 & .705 & .909 & .495 & .570\\
\hline \hline
& 24 & JOINT-BASE1
                & .641 & .611 & .794 & .779 & .961 & .479 & .547\\
& 25 & JOINT-BASE2
        & .685 & .637 & \textbf{.830} & .790 & \textbf{.991} & .515 & .597 \\
&26 &JOINT
              & \textbf{.703} & \textbf{.658} & .826 & \textbf{.805} & .969 & \textbf{.527} &\textbf{ .607}
\end{tabular}
}
\end{center}
\caption{
Label-based measures for all, head and tail types.
Largest numbers in each column for each section separated with dotted lines are in bold font.}
 \tablabel{label}
\end{table}

\tabref{label} shows the results for label-based measures 
for all  (102 types), 
head (frequency $>$
3000; 15 types) and tail (frequency $<$ 200; 36 types) types.
Each row represents one of the models. 
If not mentioned explicitly, the macro $F_1$ for all types (macro in \tabref{label})
is the measure we talk about when comparing models.

Overall, we see a similar trend as in \tabref{example}.
The only exception is that AGG-FIGER (line 23) is better here than CNN+MIML-ATT (line 22). 
JOINT (line 26) is also  worse than JOINT-BASE2 (line 25) in P@50 measures in all and head types.
The main points are still valid: (i) CNN is better for learning representations
than FF; (ii) different levels of entity representations are complementary;
(iii) ELR is very powerful; 
(iv) MIML is very effective in mitigating noise of distant supervision
and MIML-ATT is the best algorithm in this regard;
(v) CM and GM have complementary properties and their joint model performs the best.

\subsection{Analysis}

\textbf{Adding another source: description-based embeddings.}
While in this paper, we focus on the contexts and names of entities,
there is a textual source of information about entities in KBs
which we can also make use of:
descriptions of entities. 
We extract Wikipedia descriptions of \textsc{figment} entities 
filtering out the entities without description ($\sim$40,000 out of $\sim$200,000).

We then build a simple entity representation 
by averaging the embeddings of the top $k$  words (w.r.t. tf-idf)  of the description (henceforth, AVG-DES).\footnote{$k$ = 20 gives the best results on dev.}
This representation is used similar to our embeddings in \secref{gm}
to train a GM.
We also train our best GM model \linebreak ELR+SWLR+CLR(CNN)
and our best CM model CNN+MIML-ATT, as well as 
a combination (concatenation) of GM and AVG-DESC (GM+ACG-DES), 
a joint model of CM and GM (CM+GM), and finally a joint
model of all three models (GM+CM+AVG-DES) on this smaller dataset.
Since the descriptions are coming from Wikipedia, we
use 100-dimensional Glove \cite{glove14}
embeddings pretrained on Wikipdia+Gigaword 
to get a good coverage of words.
For our CM and GM models, we still use the embeddings we trained before.

Results are shown in \tabref{desc}.  We only show the micro $F_1$, the most represenative measure.
GM works better than AVG-DES, again showing the power of \gmodel{}s in this task.
The joint model, GM+CM, again improves over each single model.
A more important result is that adding AVG-DESC to GM is very effective, especially for the tail entities for which the micro $F_1$ is improved by 9\%.
This suggests that for tail entities, the contextual and name information
is not enough by itself and some keywords from descriptions can be really helpful.
Integrating more complex description-based embeddings, e.g., by using CNN \cite{xie16dkrl}, 
may improve the results further. We leave this for future work.
The best model here is the joint model (GM+CM+AVG-DES), 
with slight improvements over GM+AVG-DES.

\begin{table}[h]
\begin{center}
\begin{tabular}{l|c|c|c}
&        \multicolumn{3}{c}{entities}\\
             & all     & head     & tail \\ 
\hline
AVG-DES             &.760 & .786 & .722 \\
GM 					&.817 & .842 & .741 \\
CM 				 	&.759 & .755 & .632 \\
GM+CM				&.827 & .854 & .748 \\
GM+AVG-DES  
				    &  {.858} & {.870}  & {.830}\\
GM$+$CM+AVG-DES
					& \textbf{.865} & \textbf{.878} & \textbf{.834}

\end{tabular} 
\end{center}
\caption{Micro average $F_1$ results on the dataset of entities with Wikipedia description.}
\tablabel{desc}
\end{table}

\textbf{Comparison of character/word-level GM models on unknown vs.\ known entities.}
To do a more fine-grained comparison between
GM models that are based on an entity's name and the feature-based baselines,
we do a further analysis.
We divide  test entities into 
\emph{known entities} -- at least one word of the entity's name
appears in a train entity -- and \emph{unknown entities} (the complement).
There are 45,000 (resp.\ 15,000) known (resp.\ unknown) test entities.

\tabref{unknown}  shows that
NSL works worse than CNN on  known entities (1.2\%)
but it is much worse 
on unknown entities (by 
5.8\%), justifying our preference for deep learning CLR models.
As expected, BOW works 
relatively well for known entities and
really poorly for unknown entities. 
Basically, BOW has no clue 
for the unknown entities.
For WWLR and SWLR in our setup,
word embeddings are induced on the entire corpus using an unsupervised 
algorithm.
Thus, even for many words that
did not occur in
train, they have access to informative representations of words.
SWLR does not have OOVs and therefore works better than
WWLR on the unknown entities.
For the known entities, WWLR works better.

\begin{table}[h]
\begin{center}
\begin{tabular}{l|ccc}
& all     & known     & unknown \\ 
\hline
CLR(NSL)        & .447 & .485  & .309\\ 
CLR(CNN)        & \textbf{.471} & \textbf{.497}  & \textbf{.367}\\ 
\hdashline
BOW             & .311 & .470  & .093 \\
WWLR           & \textbf{.553} & \textbf{.581} & .424 \\
SWLR            & .552 & .574  & \textbf{.468}\\
\end{tabular}
\caption{Micro $F_1$ on the test entities of character and word-level models for
  all, known and unknown entities.
}
\tablabel{unknown}
\end{center}
\end{table}

\textbf{Assumptions that result in errors.} 
The performance of all models suffers from a number of
assumptions we made in our training / evaluation setup that 
are only approximately true.

The first assumption is that FACC1, the entity annotations 
in Clueweb corpus, is correct. However, as mentioned before, it has
a precision of only 80-85\% 
and this causes errors.
An example is the lunar crater
``Buffon'' in Freebase, a ``location''. 
Its predicted type is ``author'' because 
some FACC1 annotations of the crater link it to the Italian goalkeeper.

The second assumption of our evaluation setup is the completeness of Freebase.
There are 21,378 entities with type ``person'' in the test set and among them 2,600 entities (12\%) do not have any finer grained type.

This is potentially due to the incompleteness of entity
types in Freebase, because if a person does not have a more
fine-grained type, she/he is probably not sufficiently famous  to be in Freebase.
To confirm this hypothesis, we examine the output of \textsc{figment} on this subset of entities. 
For 62\% of the errors, the top predicted type is a subtype of person:
``author'', ``artist'' etc.  We manually typed a random
subset of 50 and found that the
predicted type is actually correct for 44 of these entities, but missing in Freebase.

The last assumption is the mapping from Freebase to FIGER.
Some common Freebase types like ``award-winner'' are not
mapped.  This negatively affects evaluation measures for
many entities. On the other hand, the resulting types do not
have a balanced number of instances. Based on our training
entities, 11 types (e.g., ``law'') have
less than 50 instances while 26 types (e.g., 
``software'') have more than 1000 instances.  Even
sampling the contexts could not resolve this problem and this led to
low performance on tail types.


\section{Conclusion}
We presented \textsc{figment}, 
a corpus-level system
that uses textual  information for fine-grained entity typing.
We designed two scoring models for pairs of entities and types: 
a \gmodel{} that computes type scores based
on aggregated entity information 
and a \cmodel{} that aggregates the type scores of individual entity contexts.
We used embeddings of characters, words, entities and types to 
represent entity names and contexts.
Our experimental results showed that the global and the context models provide
complementary information for entity typing. 
As a result, a joint model performed the best.

\section*{Acknowledgments}
We would like to thank the anonymous reviewers for their
great comments.
This work was partially supported by the European Research
Council: ERC Advanced Grant \#740516 NonSequeToR.

\section*{Appendix A. Hyperparameters}
\begin{table}[tbhp]
\begin{center}
{
\scriptsize
\begin{tabular}{l | l  c}
& model  & hyperparameters\\ 
\hline 
\multirow{12}{*}{\small GM} &
CLR(FF)      & $d_{c}=15, h_{mlp}=600$   \\
\cdashline{2-3}
& CLR(CNN)    & $d_{c}=10, w=[1,..,8]$ \\
		   && $n=100, h_{mlp}=800$  \\   

\cdashline{2-3}
& WWLR        & $h_{mlp}=400$\\
\cdashline{2-3}
& SWLR        & $h_{mlp}=400$\\

\cdashline{2-3}
& SWLR+CLR(CNN)  
		   & $w=[1,...,7]$ \\ 
           && $d_{c}=10, n=50, h_{mlp}=700$ \\
\cdashline{2-3}
& ELR        & $h_{mlp}=400$\\
\cdashline{2-3}
& ELR+SWLR    & $h_{mlp}=600$\\
\cdashline{2-3}
& ELR+CLR(CNN)
		   & $d_{c}=10, w=[1,...,7]$\\
&&                $n=100, h_{mlp}=700$\\
\cdashline{2-3}
& ELR+SWLR+CLR 
			& $d_{c}=10, w=[1,...,7]$\\
&&                $n=50, h_{mlp}=700$\\
\cdashline{2-3}
& FF          & $h_{mlp}=500, cs=10$ \\
\cdashline{2-3}
& CNN 		& $w=[1,2,3,4], n=300, h_{mlp} = 600$ \\
\cdashline{2-3}
& AVG-DES
                & $h_{mlp}=600$ \\
\cdashline{2-3}
& GM+AVG-DES
                 & $d_{c}=10, w=[1,...,8]$\\
&&                $n=100, h_{mlp}=1500$ \\
\hline 
\multirow{2}{*}{\small CM}
& FF & $cs = 10,  h_{mlp}=500$\\
\cdashline{2-3}
& CNN & $cs = 10,  h_{mlp}=600, w = [1,2,3,4]$
\end{tabular} 
}
\caption{Hyperparameters of different models.
$w$ is the CNN filter size. 
$n$ is the number of feature maps for each filter size. 
$d_c$ is the character embedding size.
$d_h$ is the LSTM hidden state size. 
$h\dnrm{mlp}$ is the number of hidden units in the output MLP.
$cs$ is the context size.
}
\tablabel{hyp}
\end{center}
\end{table}

\vskip 0.2in
\bibliography{sample}
\bibliographystyle{theapa}

\end{document}